\documentclass{article}
\usepackage{verbatim}
\usepackage{setspace}
\usepackage{graphicx}
\usepackage[centertags]{amsmath}
\usepackage{amsfonts}
\usepackage{multirow}
\usepackage{amssymb}
\usepackage{amsmath}
\usepackage{amsthm}
\usepackage{algpseudocode}
\usepackage{algorithm}
\theoremstyle{plain}

\theoremstyle{definition}

\theoremstyle{remark}

\title{Simple one-pass algorithm for penalized linear regression with cross-validation on MapReduce}
\author{Kun Yang}
\begin{document}
  \maketitle
  \begin{abstract}
    In this paper, we propose a one-pass algorithm on MapReduce for penalized linear regression
    \[f_\lambda(\alpha, \beta) = \|Y - \alpha\mathbf{1} - X\beta\|_2^2 + p_{\lambda}(\beta)\]
    where $\alpha$ is the intercept which can be omitted depending on application; $\beta$ is the coefficients and $p_{\lambda}$ is the penalized function with penalizing parameter $\lambda$. $f_\lambda(\alpha, \beta)$ includes interesting classes such as Lasso, Ridge regression and Elastic-net. Compared to latest iterative distributed algorithms requiring multiple MapReduce jobs, our algorithm achieves huge performance improvement; moreover, our algorithm is exact compared to the approximate algorithms such as parallel stochastic gradient decent. Moreover, what our algorithm distinguishes with others is that it trains the model with cross validation to choose optimal $\lambda$ instead of user specified one.
  \end{abstract}
  \section{Introduction}
  Linear regression model has been a mainstay of statistics and machine learning in the past decades and remains one of the most important tools. Given the design matrix $X = (X_1, X_2, ..., X_p) = (x_{ij})_{n\times p}\in \mathbb{R}^{n\times p}$ and response $Y$. We fit a least square linear model by minimizing the residual sum of squares
  \begin{equation}
  \textrm{RSS}(\alpha, \beta) = (Y - \alpha\mathbf{1} - X\beta)^T(Y - \alpha\mathbf{1} - X\beta)
  \label{rss}
  \end{equation}
  There are two reasons why we are often not satisfied with \ref{rss}: i) the least square estimates often have low bias but large variance, it is especially true when some of the predictors are redundant. Prediction accuracy can sometimes be improved by shrinking or setting some coefficients to zero. By doing so, we try to strike a balance between bias and variance of the model. A typical way to do shrinkage is to add some penalty terms in RSS; ii) with a large number of predictors, we often like to determine the smallest subset that exhibit the strongest effect to enhance the interpretability of the model. Shrinkage is usually achieved by adding a penalty term in the RSS then mininizing the penalized loss function.
  \par In this paper, we propose a one-pass algorithm on MapReduce for penalized linear regression. Compared to latest iterative distributed algorithms \cite{Boyd2011} requiring multiple MapReduce jobs, our algorithm achieves huge performance improvement; moreover, our algorithm is exact compared to the approximate algorithms such as parallel stochastic gradient decent \cite{Zinkevich2010}. Moreover, what our algorithm distinguishes with others is that it trains the model with cross validation to choose optimal penalty parameter instead of user specified one.
  \section{Simple One-Pass Algorithm}
 To fit the model, we need to solve the optimization
 \begin{equation}
 \label{RSSP}
 (\alpha, \beta) = \textrm{arg}\min (Y - \alpha\mathbf{1} - X\beta)^T(Y - \alpha\mathbf{1} - X\beta) + p_\lambda(\beta)
 \end{equation}
 where $p_\lambda$ is some penalty function, popular choices are Lasso, Ridge and Elastic-net penalty and $\mathbf{1}\in \mathbb{R}^{n\times 1}$ . The columns of $X$ are standardized to eliniminate the scaling issue, i.e., the columns are first centralized then scaled to unit length
 \[X = X_cD + C\]
 where $X_c$ is the standardized matrix; $D$ is a diagonal matrix where diagonal elements are the standard deviation of each column; $C$ is the center matrix with the form $\mathbf{1}(\bar{X}_1,\bar{ X}_2, ..., \bar{X}_p)$, where $\bar{X}_i$ are the averages of $X_i$, $i = 1, 2, ..., p$.
 \par We fisrt fit the model with standardized matrix $X_c$ then transform the model back to the original scale, formally
\begin{eqnarray}
 (\hat{\alpha}, \hat{\beta}) &=& \textrm{arg}\min (Y - \hat{\alpha}\mathbf{1} - X_c\hat{\beta})^T(Y - \hat{\alpha}\mathbf{1} - X_c\hat{\beta}) + p_\lambda(\hat{\beta})\label{scale1}\\
 (\alpha, \beta) &=& (\hat{\alpha} - CD^{-1}\hat{\beta}, D^{-1}\hat{\beta})\label{scale2}
 \end{eqnarray}
 Taking the first derivative of $\alpha$ and setting it to zero, we have $\hat{\alpha} = \mathbf{1}^TY / n = \bar{Y}$ and
 \begin{eqnarray}
 &&(Y - \hat{\alpha}\mathbf{1} - X_c\hat{\beta})^T(Y - \hat{\alpha}\mathbf{1} - X_c\hat{\beta})\\
 &=& Y^TY - 2\hat{\alpha} Y^T\mathbf{1} + n\hat{\alpha}^2 - 2(Y - \hat{\alpha}\mathbf{1})^TX_c\hat{\beta} + \hat{\beta}^TX_c^TX_c\hat{\beta}\\
 &=& Y^TY - 2\hat{\alpha} Y^T\mathbf{1} + n\hat{\alpha}^2 - 2Y^TX_c\hat{\beta} + \hat{\beta}^TX_c^TX_c\hat{\beta}\\
 &=& Y^TY - n\bar{Y}^2 -2(Y^TX - n\bar{Y}(\bar{X}_1, \bar{X}_2, ..., \bar{X}_p))D^{-1}\beta + \\
 &&\beta^TD^{-1}(X^TX - n(\bar{X}_1, \bar{X}_2, ..., \bar{X}_p)^T(\bar{X}_1, \bar{X}_2, ..., \bar{X}_p))D^{-1}\beta
 \end{eqnarray}
 Below are the statistics we need to calculate in the algorithm and notice that they are all additive; moreover, unlike $(X, Y)$ which usually has billions of columns and can only be stored in distributed system, these statistics can be easily loaded into memory.
 \begin{equation}\label{stats}
 n, Y^TY, X^TY, \bar{Y}, \{\bar{X}_i\}_{i = 1}^p, X^TX
 \end{equation}
 Then $D = \textrm{diag}(X^TX)^{1/2}$ and $C = \mathbf{1}(\bar{X}_1,\bar{ X}_2, ..., \bar{X}_p)$. The full description of our algorithm is in Algorithm 1, where $k$ is the number of cross validation and the rule of thumb is to set $k = 5, 10$; $\lambda$s are the list of penalty parameters. In order to train the model with cross validation, we randomly distribute each sample to one of the data chunks; then calculate the statistics (\ref{stats}) for each chunk in the reduce phase.
\begin{algorithm}[H]
\caption{Penalized Linear Regression MapReduce Algorithm}
\begin{algorithmic}[1]
\Procedure{PenalizedLR-MR}{$X, Y, k, \lambda$s}\\
\textsc{Map Phase}
\For{each sample $(x, y)$ where $x\in \mathbb{R}^{1 \times p}$ and $y$ is a scalar.}
    \State \textbf{Generate} key = random$\{0, 1, ..., k - 1\}$
    \State \textbf{Calculate} statistics in (\ref{stats}) for $(x, y)$: statistics = $[1, x, y,  y^2, xy, x^Tx]$
    \State \textbf{Emit} (key, statistics)
\EndFor \\
\textsc{Reduce Phase}
\For{each (key, value list)}
	\State \textbf{Aggregate} the whole value list and \textbf{denote} it as chunk\_statistics
	\State \textbf{Emit} (key, chunk\_statistics)
\EndFor \\
\textsc{Cross Validation Phase}
\State $\{s_i = [n_i, X_i, Y_i, Y_i^TY_i, X_i^TY_i, X_i^TX_i]\}_{i = 1}^k$ are the chunk\_statistics from previous MapReduce job
\For{each $\lambda$ in $\lambda$s}
	\For{$i\gets 0, ..., k - 1$}
		\State \textbf{train\_data} = $\sum_{k \neq i}s_k$
		\State \textbf{test\_data} = $s_i$
		\State train the model (\ref{RSSP}) with \textbf{train\_data} and calculate the mean squared prediction error $p_i$ for \textbf{test\_data}
	\EndFor
	\State mean prediction error for $\lambda$ is: pre($\lambda$) = Average of $\{p_i\}_{i = 1}^{k - 1}$
\EndFor
\State $\lambda_{\textrm{opt}} = \textrm{arg}\min\textrm{pre}(\lambda)$
\State \textbf{data} = $\sum_{i = 1}^{k - 1}s_i$
\State \textbf{train} the model (\ref{RSSP}) with \textbf{data} and \textbf{transform} the model into original scale as in (\ref{scale1}, \ref{scale2})
\State \textbf{return} $(\alpha, \beta, \lambda_{\textrm{opt}})$ or possible the prediction erros in cross validation for each $\lambda$
\EndProcedure
\end{algorithmic}
\end{algorithm}
\subsection{The Robust Distributable Algorithm}
The key is to compute \eqref{stats}. When $n$ is large, the main naive aggregation would lead to numerical instability as well as to arithmetic overflow. Here we use a robust distributable algorithm to compute \eqref{stats}.

Given $n$ $p$-dimensional row vectors \[\{x_1, x_2, ..., x_n\}\] and 1-dimensional scalars \[\{y_1, y_2, ..., y_n\}\] we calculate \[\sum_{i = 1}^n x_i/n, \textrm{covar}(x_1, ..., x_n), \sum_{i = 1}^nx_iy_i/n, \sum_{i = 1}^ny_i/n, \sum_{i = 1}^ny_i^2/n, n\] instead to avoid numerical pitfalls. We adopt the MapReduce pseudo-code to describe the distributable algorithm that calculates statistics in \eqref{stats}.
\\
\\
For the mean, it is trivial to verify that, \small\[\textrm{Mean}(x_1, ..., x_m; x'_1, ..., x'_n) = \frac{m}{m + n}\textrm{Mean}(x_1, ..., x_m) + \frac{n}{m + n}\textrm{Mean}(x'_1, ..., x'_n)\]\normalsize
In mappers, we have
\small \begin{eqnarray}
\textrm{Mean}(x_1, x_2, ..., x_n, x_{n + 1}) &=& \frac{n}{n + 1}\textrm{Mean}(x_1, ..., x_n)+ \frac{1}{n + 1}x_{n + 1}\\
= \textrm{Mean}(x_1, ..., x_n) &+& \frac{1}{n + 1}(x_{n + 1} - \textrm{Mean}(x_1, ..., x_n) )
\end{eqnarray}
\normalsize
In combiners or reducers, we have
\small
\begin{equation}
\begin{split}
\textrm{Mean}(x_1, ..., x_m; & x'_1, ..., x'_n) = \textrm{Mean}(x_1, ..., x_m)\\
 & + (1 - \frac{m}{m + n})(\textrm{Mean}(x'_1, ..., x'_n) - \textrm{Mean}(x_1, ..., x_m))
\end{split}
\end{equation}
\normalsize
For the covariance, it can be shown that
\small\[\textrm{covar}(x_1, ..., x_n) = \frac{1}{n}\sum_{i = 1}^n(x_i - \textrm{Mean}(x_1, ..., x_n))^T(x_i - \textrm{Mean}(x_1, ..., x_n))\]\normalsize
some literature defines covariance with factor $1/(n - 1)$, (\ref{covar}) below can be modified accordingly. To calculate the covariance, it is not difficult to verify that (expand the left and right hand; then compare)
\small
\begin{equation}
\label{covar}
\begin{split}
\textrm{covar}(x_1, ..., x_m; x'_1, ..., x'_n) &= \frac{m}{m + n}\textrm{covar}(x_1, ..., x_m) \\
&+ \frac{n}{m + n}\textrm{covar}(x'_1, ..., x_n) \\
&+ \frac{n}{m + n}\frac{m}{m + n}(\bar{x'}- \bar{x})^T(\bar{x'}- \bar{x})
\end{split}
\end{equation}
\normalsize
where $\bar{x} = \textrm{Mean}(x_1, ..., x_m)$ and $\bar{x'} = \textrm{Mean}(x_1, ..., x_n)$.\\
\\
So in mapper, we have
\small
\begin{equation}
\begin{split}
&\textrm{covar}(x_1, ..., x_n, x_{n + 1}) = \frac{n}{n + 1}\textrm{covar}(x_1, ..., x_n) \\
& + \frac{n}{n + 1}\frac{1}{n + 1}(\bar{x}- x_{n + 1})^T(\bar{x}- x_{n + 1})
\end{split}
\end{equation}
\normalsize
since $\textrm{covar}(x_{n + 1}) = 0$.\\
\\
In combiner and reducer, we apply (\ref{covar}).
\\
\\
Once we have \[\sum_{i = 1}^n x_i/n, \textrm{covar}(x_1, ..., x_n), \sum_{i = 1}^nx_iy_i/n, \sum_{i = 1}^ny_i/n, \sum_{i = 1}^ny_i^2/n, n\] we can recover $\sum_{i = 1}^n x_i^Tx_i/n = X^TX/n$ easily, where $X = (x_1^T, ..., x_n^T)^T$.

\subsection{Optimization}
To train the model on \textbf{train\_data} = $\sum_{k \neq i}s_k$, we need to minimize the loss function $f(\alpha, \beta)$, where
\begin{equation}
\begin{split}
f(\alpha, \beta) &= (Y - \alpha\mathbf{1} - X\beta)^T(Y - \alpha\mathbf{1} - X\beta) + p_\lambda(\beta)\\
&= Y^TY - n\bar{Y}^2 -2(Y^TX - n\bar{Y}(\bar{X}_1, \bar{X}_2, ..., \bar{X}_p))D^{-1}\beta + \\
&\quad\beta^TD^{-1}(X^TX - n(\bar{X}_1, \bar{X}_2, ..., \bar{X}_p)^T(\bar{X}_1, \bar{X}_2, ..., \bar{X}_p))D^{-1}\beta + p_\lambda(\beta)
\end{split}
\end{equation}
which is equivalent to minimize
\begin{equation}
\begin{split}
f'(\alpha, \beta) &= \beta^TD^{-1}(X^TX - n(\bar{X}_1, \bar{X}_2, ..., \bar{X}_p)^T(\bar{X}_1, \bar{X}_2, ..., \bar{X}_p))D^{-1}\beta-\\
&\quad 2(Y^TX - n\bar{Y}(\bar{X}_1, \bar{X}_2, ..., \bar{X}_p))D^{-1}\beta + p_\lambda(\beta)
\end{split}
\end{equation}
where $f'(\alpha, \beta)$ can be constructed from \textbf{train\_data} = $\sum_{k \neq i}s_k$ and minimization of $f'$ can be solved by coordinate descent algorithm \cite{Friedman2010}.

\section{Implementation}
  The commercial version of the implementation is available at Alpine Analytics Inc\textregistered: www.alpinedatalabs.com. The open source version is submitted to Apache Mahout [ISSUE 1273]\footnote{https://issues.apache.org/jira/browse/MAHOUT-1273}.
  \section{Conclusion}
  In order to fully exploit the parallelism, the cross validation phase can be implemented in another MapReduce job. This feature is not in our current version because we notice that $p$ is at the scale of $10,000$ covering most of the real word applications and it is also a physically and financially formidable task to collect billions of observations with millions of features. For the data we analyze at Alpine Analytics Inc\textregistered, they are all below the $10,000$ scale. Hence, we are confident that our version is sufficient for most applications. How to deal with more features is our future work.
  \bibliographystyle{plain}
  \bibliography{LR}
\end{document}